# GenONet: A Generative Operator Network for High-Resolution Precipitation Nowcasting

**Mohammad Kian Golkar[1], Luciano Alves de Oliveira[1*], and Mohammad Khanjani[2]**
[1]University of Illinois Urbana-Champaign [2]Sharif University of Technology

[*]Corresponding Author: Luciano8@illinois.edu

### Abstract

High-resolution precipitation nowcasting is critical for reducing the impacts of severe weather but remains difficult because of rapid storm evolution. Deep learning models have shown great promise for this task, but their predictive skill often deteriorates over longer forecast horizons. This leads to increasingly blurry forecasts that fail to capture the complex, non-linear evolution of storm systems. In order to address these limitations, we introduce Spatio-Temporal U-DeepONet (GenONet), a novel architecture for long-range precipitation forecasting up to 3 hours, specifically designed to produce sharp and physically consistent results. GenONet's architecture pioneers the use of a Deep Operator Network (DeepONet) as a generator within a Generative Adversarial Network (GAN) framework for this task. The DeepONet learns the continuous-time dynamics of precipitation, ensuring stability over long forecast horizons. Adversial training against a spatio-temporal discriminator compels the model to produce sharp, coherent forecasts, while a physics-informed loss regularizer, derived from the Moisture Conservation Equation, improves physical plausibility in our ablation setting. Quantitative evaluations show that our model achieves consistently higher scores on most of the metrics, especially for high-intensity events and at longer lead times. Qualitatively, GenONet produces structurally coherent forecasts that maintain their integrity, whereas baseline models degrade into indistinct patterns. Finally, an ablation study confirms the benefit of this physics-informed loss, highlighting the strength of combining operator learning with adversarial training.

**Keywords** Rainfall nowcasting · U-Net · Deep Operator Networks (DeepONet) · Generative adversarial networks (GAN)

## I. Introduction

Precipitation nowcasting, the high-resolution forecasting of rainfall in the immediate future (usually 0-2 hours), is critical for reducing the effects of weather related threats like flash floods, landslides, and disruptions to transportation and agriculture [1], [2]. The social and economic stakes are enormous, as extreme weather events routinely result in disastrous losses [3]. Despite its importance, nowcasting remains challenging. The two main methods are Numerical Weather Prediction (NWP) and radar based echo extrapolation (REE) for precipitation nowcasting. While powerful for longer term forecasting, traditional NWP models often lack the timeliness and fine scale detail needed for nowcasting because of high computational demands and spin-up times [4]. As a result, REE methods, which use high-resolution radar data to predict future rain, are typically better suited to this immediate, short term task. REE itself can be categorized into traditional and deep learning-based methods. Traditional methods, like those based on optical flow [5], provide fast updates but are unable to capture the highly complicated, non-linear evolution, initiation and dissipation of storm systems [6].

To overcome these limits, deep learning has become a revolutionary influence in the REE framework [7], [8]. Deep learning models are inherently data-driven, unlike their traditional counterparts. By learning from large historical radar archives, they draw on their ability to learn the complex, non linear dynamics of weather events. This allows them to implicitly model the difficult processes of storm growth and decay that simpler methods cannot [9]. On top of this powerful potential, recent architectures such as Convolutional Neural Networks (CNNs) [10], Recurrent Neural Networks (RNNs) such as ConvLSTM [11], and U-Nets [12] have shown an impressive ability to learn intricate spatio-temporal relationships directly from radar data, leading to major improvements in prediction accuracy. Among them, the U-Net architecture from medical image segmentation has emerged as a popular backbone [13]. Its symmetric encoder-decoder architecture, enhanced by skip connections, is particularly adept at retaining the fine grained spatial information of radar imagery often lost in standard CNNs [14]. Although many adaptations such as RainNet and SmaAt-UNet have proven its effectiveness [15][16], models based on this architecture still face fundamental challenges. They might not fully overcome challenges with precipitation intensity changes, rapidly moving systems, and extreme precipitation events [17]. We hypothesize that combining operator learning, adversarial training, and physics-informed regularization will yield more stable, sharp, and physically consistent forecasts. In fact, purely data-driven DL models tend to generate blurry multi-step predictions, underestimate extreme events, and can lack physical consistency [18], [19].

A promising direction for reducing the blurriness effect is the application of GANs [20]. By training a generator to make predictions and a discriminator to distinguish them from real observations, GANs encourage the creation of sharp, perceptually realistic images that preserve high frequency details [21]. For video prediction tasks like nowcasting, advanced GANs typically utilize spatio-temporal discriminators that consider entire sequences, thus ensuring not just spatial realism within every frame but also temporal consistency across the prediction [22]. This adversarial training setup helps the limitations of pixel-wise loss functions and produces forecasts that are qualitatively better. However, though adversarial training addresses the sharpness issue, it does not necessarily solve the problem of worsening performance over long lead times. Several generative models continue to adopt a recursive, frame-by-frame prediction mechanism, thereby can lead to the accumulation and propagation of errors and resulting in a loss of structural integrity for long range predictions.

These challenges of maintaining long range temporal stability have motivated the adoption of operator learning paradigms. DeepONets have emerged as a promising framework for learning operators—mappings between infinite-dimensional function spaces—instead of learning simple mappings between fixed-dimensional vectors [23]. Rather than forecasting one frame at a time, a DeepONet can learn the continuous time operator that maps the entire input sequence of radar observations to the future state over the entire forecast horizon. This is a more inherently stable and less prone to error approach than recursive models. Its modular design, comprising a branch network for encoding input functions and a trunk network for processing query coordinates (e.g., time and space), provides great flexibility [24]. Recent studies have already started to explore enhancing DeepONets with convolutional structures for spatial data processing [25] and sequential models for time-dependent inputs [26], setting the stage for its application to complex spatio-temporal tasks like nowcasting.

we introduce a novel precipitation nowcasting model, GenONet, an architecture that adapts the operator learning framework for high resolution precipitation nowcasting up to 3 hours. To the best of our knowledge, this work represents the first successful integration of the DeepONet paradigm as the core predictive engine within a generative adversarial framework for precipitation nowcasting. GenONet is based on a DeepONet generator, using the operator learning framework to ensure temporal stability on long forecast periods. To generate sharp and realistic forecasts, this generator is trained adversarially in a GAN against a spatio-temporal discriminator. Finally, to explore the potential benefits of incorporating explicit physical constraints which is a common challenge for purely data-driven generative models, we conduct a targeted ablation study. This study adds a physics-informed loss, derived from the Moisture Conservation Equation (MCE), as a regularizer to the generator's objective function to evaluate its effect on physical plausibility and forecast quality.

Our objective is to develop and evaluate GenONet for high resolution precipitation forecasting up to 3 hours. We demonstrate that our GenONet achieves advanced performance, producing sharper and more temporally consistent predictions than baseline models. The ablation studies also confirm that incorporating the physics-informed loss can provide additional improvements in predictive skill, particularly at longer lead times. This paper introduces an operator-learning approach for nowcasting and provides a clear view on the synergistic potential of merging advanced architectures with adversarial training and explicit physical laws.

## II. Related Work

In this section, we briefly describe some of the works which are related to the method that we used in this study.

### A. CNN, RNN- and U-Net-Based Sequence Models

Early deep learning studies focused on extending CNNs to the temporal domain. Shi et al. [11] introduced ConvLSTM, replacing fully connected gates with convolutional ones to capture both spatial and temporal dependencies simultaneously. Subsequent variants (e.g., ConvGRU [27], PredRNN [28], PredRNN++ [29], and E3D-LSTM [30], RainPredRNN [31], and TISE-LSTM [32]) augmented recurrent states with gradient highway units, causal LSTMs, or 3-D convolutions to mitigate vanishing gradients and model longer sequences. While these models excel at capturing local spatio-temporal correlations, their recursive, frame-by-frame prediction tends to accumulate errors, leading to blur at longer lead times [8]. To preserve fine-grained details, the U-Net encoder–decoder architecture has been widely adopted. RainNet [15] employed a multi scale U-Net to predict German radar composites, while SmaAt-UNet [16] reduced parameters by integrating depthwise separable convolutions and a CBAM attention module. Further U-Net refinements have also improved radar nowcasting skill [35]. Nested variants such as UNet++ [33] and U$^2$Net [34] further exploited multi-scale skip connections for sharper echo boundaries. Guo et al. [36] propose 3D-UNet-LSTM, an extractor forecaster framework where a 3-D U-Net first harvests volumetric spatio-temporal features and an unshared parameter ConvLSTM forecaster then produces each future frame independently. Despite these enhancements, purely data-driven U-Nets still under-represent extreme intensities and exhibit loss of high frequency detail over time.

### B. Generative Adversarial Networks for Realistic Nowcasting

To address the blurriness and unrealistic results in deterministic forecasts, a widely used approach for sharper, more realistic outputs, GAN, have become a suitable approach. GANs use a minimax game between a generator, which generates synthetic data, and a discriminator, which attempts to differentiate synthetic data from real data. This adversarial process forces the generator to learn the underlying data distribution, producing outputs that are not only pixel-wise correct but also perceptually realistic and sharp. Additionally, they can simulate a wide range of potential outcomes from the same input and improve sharpness but do not provide uncertainty. Initial applications in nowcasting proved the potential of this paradigm [37]. For instance, Tian et al. (2020) [38] introduced the GA-ConvGRU, in which a ConvGRU-based generator was trained against a standard CNN discriminator. This model produced more realistic forecasts than its non-adversarial equivalent, especially for higher intensities of rainfall. Building on this, the Deep Generative Model of Radar (DGMR) from Ravuri et al. [2] presented an architecture involving two distinct discriminators. This dual-discriminator strategy, along with techniques such as spectral normalization, enables DGMR to generate strong probabilistic forecasts that preserves fine-scale features over time. However, GAN-only approaches still struggle with stability at long horizons and lack explicit physical constraints. Recent research has further advanced the GAN framework for nowcasting. Innovations include task-segmented GANs such as TSRainGAN for high resolution heavy rainfall prediction [39], and hybrid models like ConvLSTM-TransGAN, which combine recurrent networks with Transformer based GANs to predict extreme weather [22]. Attention mechanisms have also been a prominent focus. GA-SmaAt-GNet [40] incorporated attention inside each encoder stream, while Ashesh et al. [41] uniquely placed attention within the discriminator itself to better resolve the authenticity of synthetic images. Similarly, Xu et al. [42] adopt a two stage UA-GAN that pre predicts with TrajGRU and refines with a U-Net GAN, demonstrating that cascading recurrent and convolutional stages can jointly boost accuracy and perceptual sharpness. These methods enhance realism, but they do not address operator-level temporal stability. Our work builds on the success of the spatio-temporal discriminator, using it to guide a novel generator architecture founded on the principles of operator learning.

### C. Deep Operator Networks for Scientific Machine Learning

DeepONets [23] offer a distinct paradigm by learning operators that map between infinite-dimensional function spaces rather than learning functions that map fixed-dimensional inputs to outputs. This capability is highly suitable for modeling physical systems where the input itself is a function, such as initial or boundary conditions. The canonical DeepONet architecture features a branch network to encode the input function and a trunk network to process the coordinates of the output domain (e.g., spatial and temporal locations). The outputs are then combined to generate the prediction. The flexibility of this modular structure is a key advantage. Recent research has focused on enhancing the representational power of the branch network for complex, high dimensional inputs. For data with significant spatial structure, U-Nets have been successfully integrated as the branch network [25]. Fourier Neural Operators (FNO) [43] target operator learning and have been applied to radar echo extrapolation [44]; FourCastNet [45] leveraged spectral convolutions for global dependencies, yet these are primarily designed for down-sampled grids. U-DeepONet [25] and S-DeepONet [26] incorporated U-Net branches or sequential encoders to handle complex inputs, but they still rely on L2 losses and produce overly smooth outputs. Collectively, these studies establish operator learning as a robust backbone for long-horizon prediction. However, they have not yet been leveraged as generators within an adversarial framework, which is critical for producing sharp, realistic forecasts. Our work addresses this key research gap. GenONet is, to our knowledge, the first model to successfully employ a DeepONet as the generator in a GAN for this application, extending the operator learning paradigm by incorporating adversarial training and physics-informed regularization to simultaneously enhance sharpness and physical consistency.

## III. Methodology

This study introduces a novel deep learning architecture, the Spatio-Temporal U-DeepONet named GenONet, designed for high resolution, deterministic precipitation nowcasting. The proposed model is centered on DeepONet paradigm, which is specifically adapted to learn the complex spatio-temporal dynamics of precipitation from radar data. The core predictive engine is a U-DeepONet generator that leverages a 3D-CNN-enhanced U-Net as its branch network to process input radar sequences and an MLP-based trunk network to handle query coordinates, learning the operator that maps past observations to future precipitation fields. To generate sharp and realistic forecasts, this operator network is trained adversarially as a generator within a GAN framework, guided by a spatio-temporal 3D discriminator and a composite reconstruction loss. Furthermore, in a secondary set of experiments, we investigate the impact of adding a physics-informed regularizer, derived from the MCE to the generator's loss function to promote physical consistency.

This section details the problem formulation for our two experimental tasks, the data sources and preprocessing steps, the architecture of the GenONet and its adversarial components, and the formulation of the loss functions.

### A. Problem Definition

Precipitation nowcasting is framed as a spatio-temporal sequence-to-sequence forecasting task. Let $X_t \in \mathbb{R}^{(H\times W)}$ represent a 2D radar reflectivity map at time *t*, where $X$ and $W$ are the spatial dimensions of the domain. The goal is to predict a deterministic sequence of $L_{out}$ future precipitation fields, given an input sequence of $L_{in}$ historical radar observations, $X_{out} = \{\hat{X}_{t+1}, \dots, \hat{X}_{t+L_{out}}\}$, where $\Delta T$ is the time interval between frames. To comprehensively evaluate the proposed architecture and the impact of physics-informed regularization, this study addresses two distinct nowcasting tasks. The primary task focuses on high-resolution, long-horizon forecasting, where the model uses a time interval of $\Delta T = 5$ minutes, with an input sequence length of 15 ($L_{in} = 15$) to predict an output sequence of 36 ($L_{out} = 36$). A secondary task is defined specifically to facilitate the integration and evaluation of the physics-based regularizer, which relies on auxiliary data available at a lower temporal resolution. Therefore, for this task, we align the model's time interval to $\Delta T = 30$ minutes, taking an input of 3 frames ($L_{in} = 3$) to predict an output of 6 frames ($L_{out} = 6$), covering a 180-minute forecast horizon.

### B. Data Source and Preprocessing

The main input to our model is radar reflectivity data from the Royal Netherlands Meteorological Institute (KNMI) [46]. The dataset contains 5-minute interval national radar composites for the Netherlands for the years 2008-2018. The data is spatially cropped to a 256×256 pixel area, which corresponds to a physical area of 256 km×256 km at a 1 km/pixel resolution. To prevent temporal leakage between highly correlated neighboring sequences, the dataset is split chronologically by year rather than randomly: the years 2008–2014 are used for training, 2015–2016 for validation, and 2017–2018 for testing. This year based partitioning ensures that all test sequences post-date the training period and share no temporal overlap with it, providing a stricter and more realistic evaluation of forecast skill. Precipitation rates (mm/h) are calculated from reflectivity and normalized by dividing by the maximum rainfall rate found in the training set. Since precipitation is not a common event, the data has numerous images with minimal or no rain. Accurate prediction of extreme precipitation events is critical for nowcasting, so we use a targeted data selection approach aimed at capturing both long-duration rainfall and short lived convective bursts. For every possible sequence, consisting of 15 input frames (previous 75 minutes) and 36 output frames (next 180 minutes) on 5-minute intervals, we applied a two-part selection criterion. First, to identify events with significant rainfall, we calculated the total precipitation over the entire 3-hour forecast period. This is done by summing the precipitation rates over all 36 future frames and multiplying by the time interval (5/60 hours). The resulting 256×256 grid of total precipitation is then split into 32×32 pixel blocks. If the mean total precipitation within any block is greater than a volumetric threshold (e.g., 5.0 mm over 3 hours), the sequence is chosen. This condition ensures the inclusion of widespread or long-duration rainfall events that are hydrologically important. Second, to capture short lived but intense convective events, we analyze each of the 36 future frames individually. For each frame, we again split the grid into 32×32 blocks, and if the mean precipitation rate within any block in any single future frame is greater than a high-intensity threshold (e.g., 2.0 mm/h), the sequence is also chosen. This dual-pronged strategy helps the model's training towards meteorologically significant scenarios. It makes efficient use of computational resources while making sure the training data includes both long lasting heavy rain and short, intense storms common in extreme weather. In the results reported here, no such filtering is applied: all models are trained and evaluated on the full, unfiltered sequence distribution to preserve the natural precipitation climatology and to keep our categorical scores directly comparable with prior work. For the secondary task of exploring the physics-informed regularizer, auxiliary meteorological data are necessary. To inform the physics-based component of our loss function, we use meteorological variables from the ERA5 global reanalysis dataset produced by ECMWF [47]. These variables include air temperature, wind speed components (u and v), dew point temperature, and surface net solar radiation, all of which are important in estimating terms in the MCE. ERA5 data are available at an hourly temporal resolution and on a 0.25° × 0.25° spatial grid. To align with the 30-minute temporal resolution of our secondary experimental setup, specific preprocessing steps are applied. First, cubic spline interpolation is used to upsample the hourly ERA5 variables to 30-minute intervals. Second, to match the 1 km spatial resolution of the radar data, Kriging interpolation, as implemented in the PyKrige package [48], [49], is used. Kriging is a geostatistical interpolation method that takes into account the spatial autocorrelation structure of the data, allowing the creation of smooth, continuous meteorological fields at the target 1 km resolution without imposing artificial fine scale structure. To confirm that the physics regularizer does not depend on this choice, we repeated the physics experiment using bilinear interpolation in place of Kriging and obtained comparable skill (Section IV), indicating low sensitivity to the interpolation method. This careful preprocessing ensures that the auxiliary physical variables are spatio-temporally aligned with the radar inputs and forecast outputs and provides consistent inputs to our physics-informed framework.

## C. Proposed Model Architecture

In the following, we present our proposed GenONet architecture in more detail. Since our model aims to benefit from the operator learning paradigm, advanced spatio-temporal feature extraction, and a generative adversarial framework, we first introduce its components separately to provide a deeper understanding of and reasoning for the chosen components.

**1) Deep Operator Networks.** DeepONet was originally proposed for learning operators—mappings between function spaces—rather than learning fixed-dimensional functions [23]. An operator $\mathcal{G}$ maps an input function $v$ to an output function $\mathcal{G}(v)$. DeepONet approximates $\mathcal{G}$ with two sub-networks: a branch network that encodes $v$, and a trunk network that ingests a query coordinate $t'$.The approximation reads:

(1)

$$\mathcal{G}(v)(t') \approx \sum_{k=1}^{p} b_k(v) t_k(t')$$

Where $\{b_k\}_{k=1}^{p}$ and $\{t_k\}_{k=1}^{p}$ are the outputs of the branch and trunk networks, respectively, and $p$ is the number of learned basis functions. This modular structure is highly effective for scientific machine learning tasks, as it allows for specialized architectures for the branch and trunk. In our GenONet model, the branch network is designed to process the sequence of historical radar images, while the trunk network processes the future time coordinates for which a forecast is required.

**2) Generative Adversarial Networks.** To enforce a closer agreement of the generated data with the ground truth and combat the tendency of models to produce overly smooth precipitation patterns, we employ a GAN framework. A GAN consists of a generative network *G* (the generator) and a discriminative network *D* (the discriminator) [20]. The discriminator's goal is to distinguish between real data samples (ground truth) and synthetic data produced by the generator, aiming to output values near 1 for real samples and near 0 for generated ones. The generator, in turn, is trained to produce data that is realistic enough to fool the discriminator. This dynamic results in a minimax game where the GAN applies a binary cross-entropy loss as the objective function:

$$min_G max_D \mathcal{L}_{GAN}(G, D) = \mathbb{E}_{X_{out}}[log D(X_{out})] + \mathbb{E}_{\hat{X}_{out}}[\log\left(1 - D(\hat{X}_{out})\right)] \quad (2)$$

Here, $X_{out}$ is a sequence of real radar observations drawn from the ground truth data distribution, and $\hat{X}_{out}$ is a forecast sequence produced by the generator *G*, conditioned on a historical input sequence. Since the generator and discriminator are trained adversarially, the generator is encouraged to create predictions that share the same statistical properties as the ground truth. This is particularly useful for generating realistic precipitation forecasts that should exhibit the high spatial variability and sharp gradients observed in real radar data.

**3) The Spatio-Temporal U-DeepONet Generator.** To combine the merits of the operator learning paradigm with powerful spatio-temporal feature extraction, we set up the generator *G* as a Spatio-Temporal U-DeepONet, the architecture of which is detailed in Fig. 1. The generator is composed of a specialized branch network, a trunk network, and a final projection layer. Following the DeepONet framework, the generator learns the operator that maps the input radar sequence function $v = \{X_{t-(L_{in}-1)\Delta T}, \dots, X_t\}$ to the future precipitation field at a given time coordinate $t'$.

The branch network is designed to encode the input radar sequence $v$ into a single, comprehensive latent feature map (see Fig. 2). This process begins by lifting the input sequence to a higher dimensional representation. However, unlike simpler DeepONet branches, ours first employs a 3D CNN frontend to explicitly capture spatio-temporal dynamics. The output of this stage is then passed through a temporal merging layer, which uses a 2D convolution to collapse the temporal dimension. This yields a single latent representation. A residual connection combines this with the initial features to form a latent representation, which is then processed by the deep U-Net backbone to produce the final branch output. This is significant depth allows the network to capture a rich hierarchy of spatial features, from large scale structures to fine-grained details. The U-Net architecture is composed of a symmetric encoder-decoder path. Each block in the encoder path consists of two sequential $3 \times 3$ 2D convolutional layers, each followed by Batch Normalization and a ReLU activation function. To enhance feature representation, a Squeeze-and-Excitation (SE) attention block is applied after each encoder block to adaptively recalibrate channel-wise feature responses. Downsampling is achieved via $2 \times 2$ max-pooling operations. The decoder path progressively upsamples the feature maps using $2 \times 2$ transpose convolutions and concatenates them with the high-resolution features from the corresponding encoder blocks via skip connections. This process ensures that crucial fine-grained spatial details are preserved in the final output. The trunk network, depicted in Fig. 3, is designed to generate a set of basis functions over the forecast horizon. It takes as input the temporal query coordinates $t'$, which correspond to the discrete future time steps $\{t+1, \dots, t+L_{out}\}$. These indices are first converted into continuous vector representations using sinusoidal positional encoding and then processed by a multi layer perceptron using SiLU activation to produce a unique feature vector for each future time step. The model operator, $G(v)(t')$, is then defined by the inner product of the branch and trunk network outputs. This operation combines the spatial information from the branch with the temporal information from the trunk.

This combined tensor is then mapped to the final single-channel precipitation field $\hat{X}_{t'}$ via a projection head, $P_{head}$, which is a shallow fully connected neural network $G$. The final prediction is thus given by:

$$\hat{X}_{t'} = P_{head}(G(v)(t')) \tag{3}$$

This process is repeated for all future time steps to generate the complete forecast sequence.

**4) The Spatio-Temporal Discriminator.** As shown in Fig. 4, the discriminator D is a deep 3D CNN designed to distinguish real video clips from generated ones. It is composed of a series of 3D convolutional blocks with a kernel size of $3 \times 3 \times 3$ and a stride of (1, 2, 2). Each block typically consists of a 3D convolution, a LeakyReLU activation function, and, after the first layer, 3D Batch Normalization to stabilize training. This configuration downsamples the spatial dimensions at each layer while maintaining the temporal resolution, allowing it to effectively assess both the spatial realism of individual frames and the temporal coherence of their evolution. Rather than providing a single score for the entire sequence, it operates as a PatchGAN, outputting a 3D grid of scores that evaluates the authenticity of local spatio-temporal patches. For training, we randomly sample a short clip of frames from both the real and generated sequences to serve as inputs for the discriminator.

## D. Loss Functions

For our primary high-resolution nowcasting task, which is illustrated in Fig. 5, the generator is trained to minimize an objective function that combines a reconstruction loss $L_{recon}$, an adversarial loss $L_{GAN}$. The reconstruction loss measures pixel-wise dissimilarity between the generated forecast sequence $\hat{X}_{out}$ and the ground truth sequence $X_{out}$. While pixel-wise losses such as $L_1$ are common for ensuring intensity accuracy, they have a known tendency to produce overly smooth or blurry forecasts, as they do not account for the perceptual quality and structural integrity of the image [50]. It is a weighted sum of a Multi-Scale Structural Similarity Index (MS-SSIM) loss and a weighted $L_1$ loss:

$$L_{recon} = L_{MS-SSIM} + \beta \cdot L_{1,weighted} \tag{4}$$

$$L_{MS-SSIM} = 1 - MS - SSIM\ (\hat{X}_{out}, X_{out}) \tag{5}$$

This composite approach leverages the complementary strengths of each component: the $L_1$ loss enforces pixel-level accuracy in intensity, while the MS-SSIM loss preserves the structural contrast and texture of precipitation fields.

This combination has been shown to improve overall forecast quality in nowcasting tasks by balancing pixel accuracy with perceptual realism [50].

In this paper, rather than embedding physics directly into the main architecture or the discriminator, we investigate the effect of adding a physics-informed loss derived from the MCE directly as a regularizer to our DeepONet generator. This allows us to systematically evaluate its impact on forecast stability in a targeted ablation study. To investigate the impact of physical constraints in our secondary experiment, we introduce an additional physics-informed loss as a regularization term. The complete model architecture for this secondary experiment, illustrating the adversarial loop and the integration of the physics-informed loss, is shown in Fig. 6. To enhance the physical plausibility and meteorological consistency of the forecasts, $L_{phys}$ introduces a soft constraint based on the principles of atmospheric moisture conservation. The MCE is a fundamental equation in atmospheric science that describes the budget of atmospheric water vapor.

A general form of the MCE can be written as:

$$\frac{\partial q}{\partial t} = -V \cdot \nabla q - \omega \frac{\partial q}{\partial p} + E - C \tag{6}$$

where $q$ is specific humidity, $V$ is the horizontal wind vector, $\omega$ is the vertical velocity in pressure coordinates, $E$ is evapotranspiration, and $C$ is condensation (related to precipitation $P$).

Directly incorporating the full 3D MCE is challenging due to the difficulty in obtaining accurate, high resolution measurements of vertical wind speed $\omega$ and the vertical profile of specific humidity $\frac{\partial q}{\partial z}$ at the scales relevant to nowcasting. Omitting the vertical advection term simplifies the equation but risks neglecting important vertical moisture transport dynamics. To test the significance of this term, we additionally implemented the vertical moisture transport $\omega \cdot \frac{\partial q}{\partial p}$ using ERA5 pressure level vertical velocity and specific humidity at 1000, 925, 850, and 700 hPa, taken as the column mean across these layers, and evaluated the extended formulation. As reported in Table V, including the vertical term did not improve skill and slightly reduced the categorical scores, which we attribute to the weak and noisy near surface vertical velocity at the radar scale and the coarse vertical resolution of ERA5. This result supports the use of the simplified 2D horizontal formulation for the present task.

To address this pragmatically while still incorporating key physical processes, we adapt a simplified 2D horizontal MCE. This approach, which simplifies the MCE to focus on multi-level horizontal moisture advection, is inspired by recent work in physics-informed nowcasting, such as the PID-GAN framework proposed by Yin et al [19]. This simplification focuses on horizontal moisture advection at multiple lower-atmospheric levels, using available ERA5 reanalysis data for 10-meter and 100-meter winds $(u_{10}, v_{10}, u_{100}, v_{100})$. To represent the net advective effect, the horizontal wind components are combined, for example by averaging, to create an effective wind field $(u_{avg}, v_{avg})$. This approach partially accounts for variations in atmospheric moisture transport without requiring explicit vertical wind data. The residual of this simplified MCE, $R_q$, which should approach zero for a physically consistent forecast, is defined at each pixel $(y, x)$ for each predicted mean precipitation frame $\hat{X}(y, x, t')$ as:

$$R_q(y, x, t') = \left(\frac{\partial q}{\partial t'} + \alpha\left(u_{avg}\frac{\partial q}{\partial x} + v_{avg}\frac{\partial q}{\partial y}\right)\right) - ET + C_{pred} \tag{7}$$

In this formulation, $q$ represents specific humidity, which is derived from ERA5 2-meter temperature, 2-meter dew point temperature, and surface pressure through the Clausius Clapeyron relation. The auxiliary variables are the effective horizontal wind components, the evapotranspiration $ET$, and $q$, all derived from the preprocessed ERA5 datasets. Here $ET$ is the Makkink reference evapotranspiration,

$$ET = 0.65 \cdot \frac{\Delta}{\Delta+\gamma} \cdot \frac{R_S}{\lambda} \tag{8}$$

where $\Delta$ is the slope of the saturation vapor pressure curve, $\gamma$ is the psychrometric constant, $\lambda$ is the latent heat of vaporization, and $R_s$ is the ERA5 incoming solar radiation. The condensation sink $C_{pred}$ is derived from the predicted precipitation rate P (mm/h) through the unit conversion $C_{pred}$ = P · $\rho_w$/ (3.6 × $10^6$ · $\rho_a$ · Δz), with $\rho_w$ = 1000 kg $m^{-3}$, $\rho_a$ = 1.2 kg $m^{-3}$, and column depth Δz = 2000 m, where the constant 3.6 × $10^6$ converts P from mm/h to m $s^{-1}$. This yields $C_{pred}$ in $s^{-1}$ (approximately 1.16 × $10^{-7}$ $s^{-1}$ per mm/h). Spatial derivatives $\left(\frac{\partial q}{\partial x}, \frac{\partial q}{\partial y}\right)$ are approximated using central differences, and the temporal derivative $\frac{\partial q}{\partial t'}$ is approximated using finite differences between the interpolated $q$ values at consecutive 30 minute forecast steps. The physics-informed loss term, is then calculated as the mean squared error of this MCE residual, averaged over all predicted frames and spatial grid points:

$$L_{phys} = \left(\frac{1}{L_{out}HW}\right) \sum_{j=1}^{L_{out}} \sum_{y,x} \left[R_q(y, x, t'_j)\right]^2 \tag{9}$$

where $t'_j$ represents the $j_{th}$ future time step. A scaling factor, α, is applied to the advection terms to bring them to a numerical range comparable to the source and sink terms; α is selected on the validation set, and the sensitivity of forecast skill to the physics weight is characterized in the ablation study (Table IV). The hyperparameter $\lambda_{phys}$ balances the influence of this physics-based regularization against the other loss terms. By penalizing deviations from this MCE approximation, $L_{out}$ encourages the model to learn forecasts that are not only accurate with respect to the observed radar data but also more consistent with the underlying principles of atmospheric water balance.

The complete loss for this ablation study is:

$$L_G = L_{GAN} + \lambda_{recon} \cdot L_{recon} + \lambda_{phys} \cdot L_{phys} \tag{10}$$

where $\lambda_{recon}$ and $\lambda_{phys}$ are hyperparameters that balance the contribution of each loss component.

### E. Experimental Setup

All models were set up in the PyTorch deep learning framework and trained on a high performance computing cluster equipped with an NVIDIA A100 GPU. Both the training and evaluation of the primary nowcasting task and the secondary physics-informed ablation study were conducted according to a rigorous protocol to provide robustness and reproducibility. The generator and the discriminator were optimized separately with the AdamW optimizer. According to empirical tuning, different learning rates were assigned: 1e-4 for the generator and 4e-4 for the discriminator, both with $\beta$ values of (0.5, 0.999). A weight decay of 0.0001 was added to the generator's optimizer to facilitate regularization.

To control the learning process dynamically, an adaptive ReduceLROnPlateau scheduler was used for each optimizer. This scheduler tracks the validation reconstruction loss and decreases the respective learning rate by a factor of 0.5 when no improvement is seen over 3 consecutive epochs. Training was done for a total of 50 epochs with a batch size of 8, a necessity imposed by the large memory footprint of the 3D spatio-temporal models. During training, the model performance on the validation set was assessed after every epoch, and the generator and discriminator weights at the lowest validation loss were saved as the final models for inference. The generator's objective function, defined in Section 3.4, is a composite loss. The adversarial term is the binary cross-entropy with logits loss. The reconstruction term is a weighted sum of a Multi-Scale Structural Similarity Index (MS-SSIM) loss and a custom weighted L1 loss. This L1 loss uses a dynamic weighting scheme to combat the imbalanced nature of precipitation data: pixels with rainfall rates in the range 0.01 to 5.0 mm/h are given a weight of 5.0, rates greater than 5.0 mm/h are given a weight of 10.0, and all other pixels are assigned a base weight of 1.0. For our main experiment, the final generator loss, where the reconstruction weight is 100 and adversarial weight is 1 to balance the adversarial and pixel-wise objectives. For the secondary ablation study, the physics-informed regularizer is appended to this objective, and its influence is governed by the hyperparameter $\lambda_{phys}$. A sensitivity analysis for this hyperparameter was performed in the section 3.3. Central architectural hyperparameters for our GenONet were determined via a combination of proven best practices and initial validation. The deep U-Net backbone of the branch network has an initial filter size of 64 channels. This number of channels doubles at each of the first four downsampling steps, up to a maximum of 1024 channels, which is kept for the next five deeper steps to trade off model capacity against computational feasibility. The MLP of the trunk network has an embedding dimension of 128 and comprises 8 linear layers. For adversarial training, the spatio-temporal discriminator takes input video clips of length 3 consecutive frames.

### F. Baseline Models and Evaluations Metrics

This study selects five representative models in the field of precipitation forecasting for comparison to assess the performance of the newly proposed network. These models are UNet, SmaAtUNet, GAN-UNet, RainNet, and DGMR, the last of which is a generative adversarial model included to benchmark GenONet against a strong generative baseline. DGMR was trained and evaluated on the KNMI data under the same chronological protocol using the canonical open source implementation, and extended to the full 180 minute horizon for a like for like comparison. The evaluation is conducted using a comprehensive suite of standard quantitative metrics. To assess the continuous fields, we use the Mean Absolute Error (MAE) for pixel-wise error and the Structural Similarity Index (SSIM) to evaluate the preservation of perceptual and structural features. For event-based evaluation, forecasts and observations are first binarized using specific rainfall intensity thresholds (e.g., 0.1, 0.5, 2 mm/h). From the resulting contingency table (Table I), we compute several key categorical metrics. The Critical Success Index (CSI), measures the fraction of correctly predicted rain events, making it effective for rare events. The Heidke Skill Score (HSS) quantifies the model's accuracy improvement over a random guess by accounting for all elements of the confusion matrix, including correct non-events. Additionally, the False Alarm Ratio (FAR) measures the proportion of predicted rain events that did not occur, indicating the model's tendency to over-forecast. Finally, to evaluate skill across various spatial scales, the Fractions Skill Score (FSS) compares the fractional coverage of rainfall within defined neighborhoods.

TABLE I
THE CONTINGENCY TABLE USED FOR CALCULATING CATEGORICAL FORECAST VERIFICATION METRICS.

| | **Predictive Positive** | **Predictive Negative** |
|---|---|---|
| **Actual Positive** | Hits (True Positive) | Misses (False Negative) |
| **Actual Negative** | False Alarms (False Positive) | Correct Negatives (True Negative) |

## IV. Results

This section provides a comprehensive assessment of the proposed GenONet architecture. We begin by comparing this model with a number of baseline models, evaluating the performance using both quantitative metrics and qualitative visualization. This is followed by a targeted ablation study to systematically examine the influence and effectiveness of the physics-informed loss regularizer.

### A. Quantitative Evaluation

A quantitative comparison is shown in Fig. 7. It plots the CSI, FAR, and HSS versus forecast lead time for three different rainfall intensity thresholds (0.1, 0.5, and 2 mm/h). Across most thresholds and lead times, GenONet (purple line) attains competitive to superior CSI and HSS, with its advantage widening at longer lead times. DGMR is competitive at short lead times but its skill degrades markedly at longer horizons and higher intensities, consistent with previous reports in the DGMR literature. Closer examination of the CSI plots reveals several important points. At the lowest threshold of 0.1 mm/h, corresponding to light precipitation, GenONet is competitive with the baselines at short lead times and maintains a higher CSI at longer lead times. This indicates superior performance in detecting the presence of rain. This performance becomes particularly marked at higher intensity thresholds. For example, in the 2 mm/h plot, the advantage of GenONet is most clear. While all models exhibit a natural degradation in performance as the forecast horizon extends, the rate of degradation for GenONet is slower, and the gap between it and the baseline models widens after the 60-minute mark. This demonstrates GenONet's strength in forecasting significant precipitation events and its robustness at longer horizons. This implies that the operator learning framework effectively captures the complex, non-linear dynamics of storm systems over time, while the adversarial training component ensures that high-intensity features are not smoothed out. Complementing its high hit rate, GenONet also maintains among the lower FAR across the forecast period, as shown in the middle column of Fig. 7. A lower FAR indicates that the model is less likely to forecast precipitation events that do not occur, an essential characteristic for trusting the operational forecasting system.

The combination of a high CSI and a low FAR indicates that GenONet attains a better balance between sensitivity (detecting true events) and precision (avoiding false alarms). The HSS plots also show a similar trend seen in the CSI, with GenONet consistently performing better than other baselines. The widening HSS gap at longer lead times, especially at the 0.5 and 2 mm/h thresholds, shows that GenONet's skill is not only higher but also more durable across the 180-minute forecast. Collectively, the findings in Fig. 7 offer strong evidence that the architectural advances in GenONet directly translate into improvements in forecast skill across on several standard evaluation metrics.

To assess forecast skill across various spatial resolutions, we use the FSS, a neighborhood-based verification metric that measures the spatial agreement between forecast and observed precipitation fields. The results for FSS are visualized as heatmaps in Fig. 8, calculated at three lead times (+30, +90, and +150 minutes), three rainfall intensity thresholds, and three spatial scales (1, 5, and 10 km). The heatmaps show that GenONet is competitive with the baselines at short lead times and shows its clearest advantage at longer lead times and higher intensities. At the shortest lead time of +30 minutes, the FSS values of GenONet are comparable to those of the baselines, reflecting a similarly accurate initial prediction of the precipitation pattern. This performance difference grows at longer lead times, where the limitations of the other models become apparent. At the +90 minute lead time, the FSS scores for the baseline models such as U-Net and RainNet show a marked decline, particularly for the 2.0 mm/h threshold. In contrast, GenONet sustains significantly higher scores, demonstrating its better ability in retaining the spatial structure for intense precipitation features over time. This strength is most evident at +150 minutes, where baseline models lose nearly all useful spatial skill at high intensities, while GenONet maintains much higher FSS values.

GenONet maintains a higher level of skill than the baselines at long lead times, with the largest relative gaps appearing at the higher intensity thresholds where the baselines approach zero. This persistence in maintaining the spatial characteristics of precipitation fields at long lead times is an advantage of the GenONet architecture. The high FSS of the model suggests its ability to produce spatially coherent and properly located precipitation fields.

Fig. 9 offers a clear comparison of pixel-wise error (MAE) and SSIM. On MAE, all models perform comparably, with the curves staying tightly grouped across the 180-minute horizon; GenONet remains competitive throughout rather than achieving the lowest error. On SSIM, GenONet retains the highest structural similarity at longer lead times, and its slower rate of decline is noteworthy, while all models degrade with lead time, GenONet better preserves the spatial structure, texture, and organization of the precipitation fields as the horizon extends. A lower MAE indicates better prediction of the exact rainfall intensity, while a higher SSIM represents better preservation of spatial structure. This structural advantage is attributable to the operator learning framework, which effectively models temporal evolution, coupled with adversarial training, which maintains high-frequency spatial details.

## B. Qualitative Analysis

Beyond quantitative metrics, a qualitative evaluation is essential to determine the model's ability to produce physically realistic and useful forecasts. Fig. 10 presents a visual comparison of the forecasts from all models for a representative case study at multiple lead times up to 180 minutes. The behavior of the baseline models illustrates typical failure modes in deep learning based nowcasting. Models trained with pixel-wise losses, including UNet, SmaAt-UNet, and RainNet, exhibit a huge loss of detail over time. Although they capture the general position of the precipitation at short lead times (e.g., T+30 min), they suffer from extreme spatial smoothing and blurriness as the forecast extends. By T+120 minutes, their outputs have changed into diffuse, indistinct patterns that fail to preserve the sharp gradients and underestimate the high intensity cores (reds and yellows), losing the coherent structure of the precipitation band. The GAN-UNet model produces forecasts that are visually sharper than the other baselines. Nevertheless, it fails to preserve temporal coherence and structural accuracy; the resulting patterns look fragmented and physically unrealistic, failing to capture the organized nature of the developing storm. DGMR produces sharp textures at short lead times but progressively loses coherence and develops artifacts as the horizon extends. In contrast, GenONet generates forecasts that are qualitatively better and close to the ground truth. It successfully captures the evolution of the elongated precipitation band, including its high-intensity leading edge. Even at T+180 minutes, GenONet preserves the band's structural integrity, accurately forecasting its shape, location, and high-intensity core.

## C. Ablation Study on Physics-Informed Regularization

This physics ablation is conducted on the secondary task ($\Delta T$ = 30 min), for which the hourly ERA5 auxiliary fields align naturally with the forecast cadence. We also applied the physics term to the primary 5 minute task; there its effect was marginal, because interpolating the hourly auxiliary fields to a 5 minute cadence smooths the moisture tendency and reduces the genuine sub hourly signal available to the physics residual. To thoroughly examine the synergistic potential of combining our generative operator learning framework with explicit physical principles, we performed a targeted ablation study on the effect of a physics-informed loss regularizer derived from the MCE. Table IV shows that the mean MCE residual decreases as the physics weight increases, from $3.63 \times 10^{-5}$ s$^{-1}$ without the constraint to $2.15 \times 10^{-5}$ s$^{-1}$ at $\lambda = 1.0$. This confirms that the physics-informed loss guides the generator toward predictions that better satisfy moisture conservation. This increased physical consistency corresponds to higher forecast skill, discussed in Table II.

TABLE II

ABLATION STUDY RESULTS COMPARING THE CATEGORICAL FORECAST SKILL OF THE STANDARD GENONET AGAINST THE VERSION ENHANCED WITH A PHYSICS-INFORMED LOSS REGULAIZER.

| Rainfall Threshold (mm/h) | Model | CSI | FAR | HSS |
|---|---|---|---|---|
| 0.1 | GenONet | 0.4817 | 0.3277 | 0.5440 |
| | GenONet + physics Loss | 0.4944 | 0.3261 | 0.5482 |
| 0.5 | GenONet | 0.3540 | 0.4945 | 0.4537 |
| | GenONet + physics Loss | 0.3599 | 0.4820 | 0.4605 |
| 2 | GenONet | 0.1606 | 0.6227 | 0.2576 |
| | GenONet + physics Loss | 0.1666 | 0.5570 | 0.2626 |

The results consistently show that the GenONet + physics Loss model produces higher CSI and HSS values and a lower FAR for all examined rainfall thresholds (0.1, 0.5, and 2 mm/h). For example, at the 0.5 mm/h threshold, the CSI improves from 0.3540 to 0.3599 and HSS from 0.4537 to 0.4589, while FAR decreases from 0.4945 to 0.4820. The benefit is most pronounced at the 2 mm/h threshold, where FAR drops markedly from 0.6227 to 0.5570 and CSI rises from 0.1606 to 0.1666. This illustrates that encouraging the model to follow physical laws gives a useful inductive bias, resulting in more precise and trustworthy event based predictions, especially for higher intensity events where physical evolution becomes more dominant.

To assess sensitivity to the ERA5 downscaling method, we retrained the physics regularized model using bilinear interpolation in place of Kriging. As shown in Table III, the two methods yield comparable skill across all thresholds, confirming that the physics regularizer does not rely on kriging induced fine scale structure.

TABLE III

SENSITIVITY OF THE PHYSICS REGULARIZED MODEL TO THE ERA5 DOWNSCALING METHOD (KRIGING VERSUS BILINEAR INTERPOLATION), SECONDARY TASK. ALL METRICS ARE COMPUTED ON PHYSICAL RAINFALL FIELDS IN MM/H FOR THE SECONDARY TASK ($\Delta$T = 30 MIN).

| Interpolation | CSI (0.1 mm/h) | CSI (0.5 mm/h) | CSI (2 mm/h) | HSS (0.5 mm/h) | FAR (0.5 mm/h) | MAE |
|---|---|---|---|---|---|---|
| Kriging | 0.4941 | 0.3525 | 0.1666 | 0.4505 | 0.4820 | 0.2617 |
| Bilinear | 0.4690 | 0.3346 | 0.1145 | 0.4391 | 0.4875 | 0.2678 |

Additionally, Table IV reports the sensitivity analysis of the hyperparameter λ, which controls the weight of the physics loss. The results reveal a nonmonotonic tradeoff between physical consistency and data driven accuracy. A moderate constraint improves categorical skill over the unconstrained baseline, with CSI at 0.1 mm/h, CSI at 0.5 mm/h, and HSS at 0.5 mm/h all peaking at $\lambda = 0.33$, while MAE and the mean MCE residual reach their minima at $\lambda = 1.0$. Beyond $\lambda = 1.0$, performance degrades rapidly, HSS at 0.5 mm/h drops to 0.2212 at $\lambda = 2.0$ and collapses to 0.0530 at $\lambda = 5.0$, while MAE rises to 0.3097 and CSI at 0.1 mm/h falls to 0.2717. Notably, the MCE residual also increases again at large λ, rising back toward the unconstrained value. This indicates that an excessive physics weight does not simply enforce the constraint more strongly at the expense of accuracy; it destabilizes training to the point that further gains in physical consistency are lost as well. Considering the intensity dependence, the physics constraint at its optimal weight improves CSI at all three thresholds (Table II), with the largest relative benefit at the highest intensity. This analysis identifies moderate weights, between approximately 0.33 and 1.0, as the effective operating range: $\lambda = 0.33$ is preferred when categorical skill is prioritized, while values toward 1.0 yield the best MAE and physical consistency. An excessive weight, by contrast, sacrifices the data driven reconstruction on which the model relies. As our dataset covers a single geographic domain, stratification across distinct regional climate regimes is left for future work, in which region adaptive weighting could be explored.

TABLE IV

SENSITIVITY ANALYSIS OF THE HYPERPARAMETER $\lambda_{phys}$, WHICH CONTROLS THE WEIGHT OF THE PHYSICS-INFORMED LOSS.

| $\boldsymbol{\lambda_{phys}}$ | MAE (mm/h) | CSI (0.1 mm/h) | CSI (0.5 mm/h) | HSS (0.5 mm/h) | Mean MCE residual ($s^{-1}$) |
|---|---|---|---|---|---|
| 0 | 0.2668 | 0.4817 | 0.3540 | 0.4537 | $3.63\times10^{-5}$ |
| 0.1 | 0.2602 | 0.4610 | 0.3385 | 0.4397 | $2.62\times10^{-5}$ |
| 0.33 | 0.2617 | 0.4941 | 0.3599 | 0.4609 | $2.59\times10^{-5}$ |
| 0.5 | 0.2598 | 0.4789 | 0.2988 | 0.4512 | $2.49\times10^{-5}$ |
| 1.0 | 0.2571 | 0.4726 | 0.2923 | 0.4294 | $2.15\times10^{-5}$ |
| 2.0 | 0.2742 | 0.3998 | 0.2855 | 0.2212 | $2.85\times10^{-5}$ |
| 5.0 | 0.3097 | 0.2717 | 0.2741 | 0.0530 | $3.00\times10^{-5}$ |

Moreover, Table V compares the horizontal only physics formulation against the extended horizontal plus vertical formulation at the optimal physics weight; the vertical term does not improve skill, supporting the 2D horizontal simplification.

TABLE V

EFFECT OF ADDING THE VERTICAL MOISTURE TRANSPORT TERM $\omega \cdot \frac{\partial q}{\partial p}$ TO THE MCE RESIDUAL, SECONDARY TASK AT THE OPTIMAL PHYSICS WEIGHT.

| Configuration | CSI (0.1 mm/h) | CSI (0.5 mm/h) | CSI (2 mm/h) | HSS (0.5 mm/h) | FAR (0.5 mm/h) | MAE |
|---|---|---|---|---|---|---|
| Horizontal | 0.4944 | 0.3599 | 0.1666 | 0.4609 | 0.4820 | 0.2617 |
| Only | | | | | | |
| Horizontal + Vertical | 0.4760 | 0.3455 | 0.1577 | 0.4435 | 0.5005 | 0.2621 |

### D. Uncertainty Quantification

Beyond deterministic forecasts, operational risk assessment benefits from an estimate of predictive uncertainty. To provide this, we equip GenONet with Monte Carlo dropout: dropout layers introduced in the generator are kept active at inference, and multiple stochastic forward passes are drawn for each input to form a predictive ensemble. From this ensemble we compute the Continuous Ranked Probability Score (CRPS), a standard probabilistic verification metric, together with the ensemble spread, as functions of lead time. As shown in Fig. 11, both CRPS and ensemble spread exhibit a pronounced local maximum near the 30 minute lead time, decrease to a minimum near 80 minutes, and then grow steadily out to 180 minutes, reaching a spread of approximately 0.035 with a mean CRPS of 0.0677 over the 3 hour horizon. The early peak occurs at the lead times where storms develop and reorganize most rapidly, so the ensemble members disagree most strongly there. After this window, the forecasts become smoother and more large scale, and the members agree more closely. At longer lead times, uncertainty grows again as forecast errors accumulate. The monotonic growth beyond 80 minutes is consistent with the expected increase of forecast uncertainty with horizon.

To place these results in context, we also evaluated DGMR, which is natively probabilistic, using the same ensemble size, test subset, and CRPS estimator. GenONet attains a substantially lower mean CRPS (0.0677 versus 0.221) with a larger ensemble spread (0.023 versus 0.015). DGMR's CRPS decreases monotonically with lead time, from 0.44 at +5 min to 0.10 at +180 min, while its ensemble members remain nearly identical. This behavior reflects the progressive weakening of its forecasts at long lead times, visible in Fig. 10, because radar fields are predominantly dry, a forecast that approaches zero attains a low absolute error without providing useful information. CRPS alone therefore does not distinguish a skillful forecast from a degenerate one on sparse fields, which is why we report it alongside the categorical and neighborhood based metrics.

## V. Discussion

Our results demonstrate that GenONet improves sharpness, temporal stability, and physical consistency compared to leading deep learning baselines. Our approach synergistically combines three key innovations: a DeepONet paradigm for modeling temporal dynamics, a GAN framework for producing sharp forecasts, and a physics-informed loss that enhances physical consistency in our ablation setting at the 30 minute cadence. This discussion synthesizes the findings, contextualizes their significance, acknowledges limitations, and outlines directions for future research.

The superior performance of GenONet can be attributed to the successful integration of its core components. First, the sharpness and realism of the forecasts, evident in the qualitative analysis (Fig. 10) and in its retention of structural similarity at longer lead times (Fig. 9), are a direct consequence of the adversarial training. The spatio-temporal 3D discriminator forces the generator to produce image sequences that are statistically indistinguishable from real radar data, effectively overcoming the inherent tendency of pixel-wise losses (like MSE) to produce blurry, averaged-out predictions. Second, the model's robustness over long forecast horizons stems from its foundation in the DeepONet framework. Instead of a simple frame-to-frame recursive prediction that accumulates errors, GenONet learns a continuous-time operator that maps the entire input radar sequence to the future state. The U-Net-based branch network extracts a comprehensive latent representation of the initial atmospheric state, while the trunk network provides a set of temporal basis functions. This architecture is inherently more stable for long-term extrapolation compared to traditional recurrent models. Third, our ablation study confirms that integrating explicit physical constraints provides a tangible benefit. While the GAN framework learns the implicit physics from the data distribution, adding the MCE-based loss as a soft constraint provides a valuable inductive bias. It guides the model towards solutions that not only fit the data but also respect a fundamental law of atmospheric science, leading to improved categorical skill scores (Table II) and demonstrably more physically consistent outputs (Fig. 10).

While promising, GenONet has limitations that must be addressed before operational use. First, the use of a deep U-Net branch and a 3D spatio-temporal discriminator is computationally intensive. We note that the discriminator is used only during adversarial training and is not part of the inference pipeline, so the deployed model already excludes its memory and compute cost; the primary target for compression is therefore the U-Net branch of the generator. Future work could apply structured pruning to remove redundant channels, knowledge distillation to train a lightweight student network from the full GenONet, and post training quantization to reduce the memory footprint, producing versions suitable for edge deployment without significant loss of accuracy. Such compression would help enable real time operational use in systems for flash flood warning and air traffic management. Moreover, the physics constraint relies on a simplified moisture budget. We tested an extension that adds the vertical moisture transport term using ERA5 pressure-level data (Table V), but it did not improve skill at the radar scale, which we attribute to the weak and noisy near-surface vertical velocity and the coarse vertical resolution of ERA5. Future research could investigate richer physical constraints through higher-resolution vertical data or by coupling the model with simplified physical simulators or NWP output to provide a stronger physical context.

## VI. Conclusion

GenONet addresses the key challenges of blurriness, physical inconsistency, and long-range degradation in precipitation nowcasting. GenONet's strong performance, validated against competitive baselines, is rooted in its unique fusion of three core components. A DeepONet provides temporal stability by learning a continuous-time operator that avoids recursive error accumulation. This is trained within a GAN framework, which enforces the generation of sharp, statistically realistic forecasts. Furthermore, a physics-informed loss based on the MCE acts as an effective regularizer, enhancing the physical plausibility and skill of the predictions in our ablation setting. Comprehensive evaluations demonstrated that GenONet produces forecasts that are sharper and more coherent than competing models, with its clearest advantage in categorical skill for higher-intensity events at longer lead times. Future work can focus on reducing computational demands and incorporating richer physical constraints. GenONet establishes a benchmark for data driven nowcasting and shows that combining operator learning, adversarial training, and physical laws is a promising path for next-generation meteorological models.

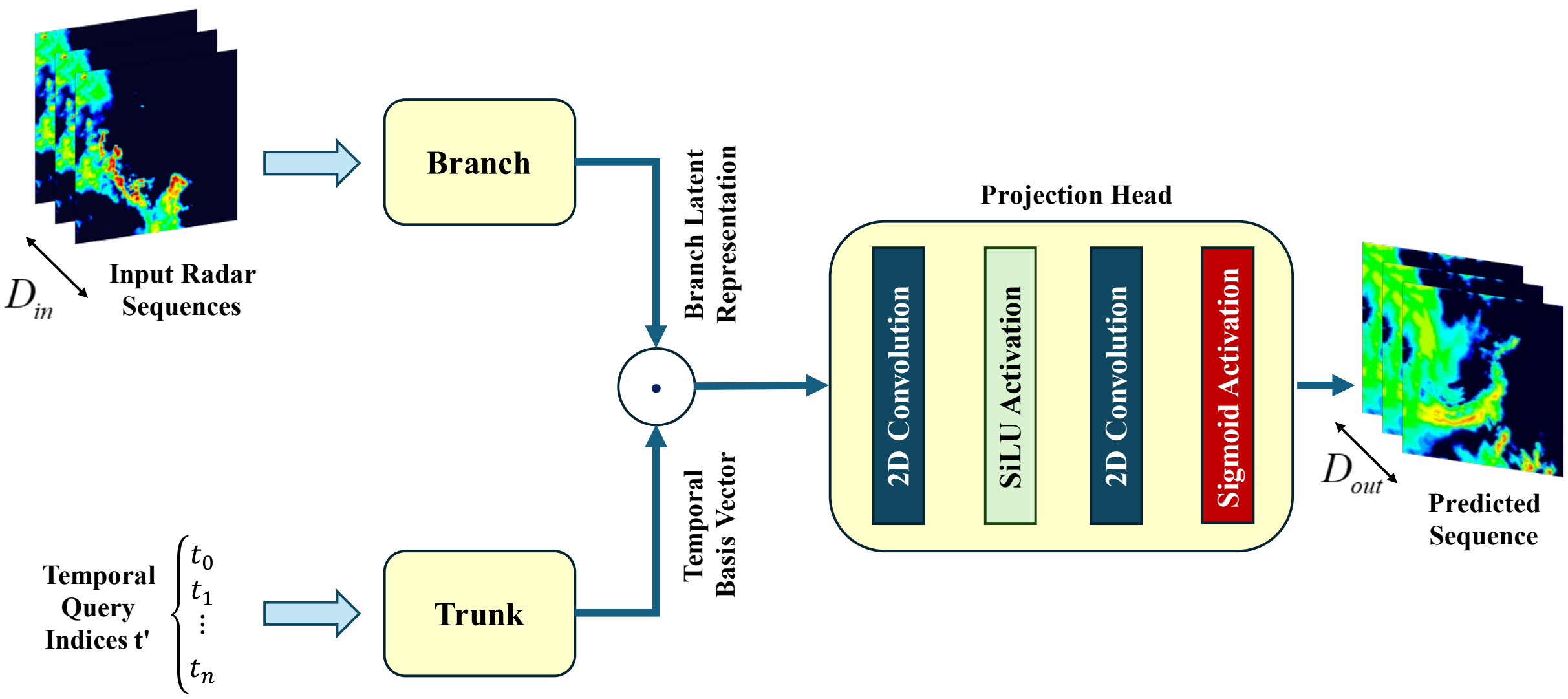

Fig. 1. Architectural diagram of the GenONet generator, which is based on DeepONet paradigm. Input radar sequences are fed into a powerful Branch network to extract a comprehensive spatio-temporal feature representation. Simultaneously, future temporal query indices ($t'$) are processed by a simpler Multi-Layer Perceptron Trunk network to generate time-dependent basis functions. The outputs of these two networks are combined and passed through a final Projection Head to produce the predicted precipitation frame for each queried time.

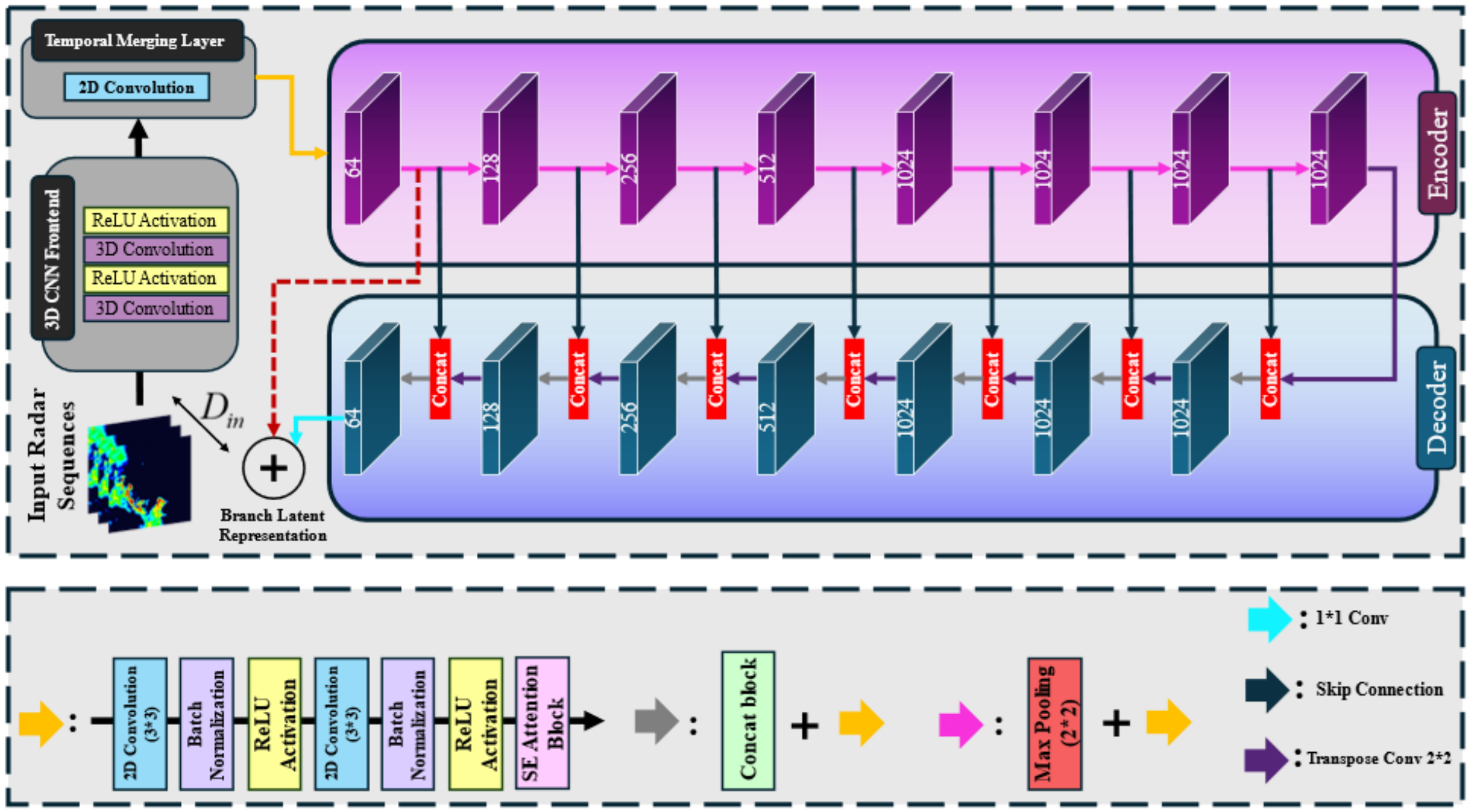

Fig. 2. Detailed architecture of the Branch Network. The network begins with a 3D CNN Frontend to capture initial spatio-temporal dynamics from the input radar sequence. A Temporal Merging Layer then collapses the time dimension, producing a single latent representation that is processed by a deep U-Net backbone. The encoder path (left) progressively extracts hierarchical features through a series of convolutions and max-pooling operations, while the decoder path (right), aided by skip connections (Concat), reconstructs the feature map, effectively preserving fine-grained spatial details crucial for sharp precipitation forecasts.

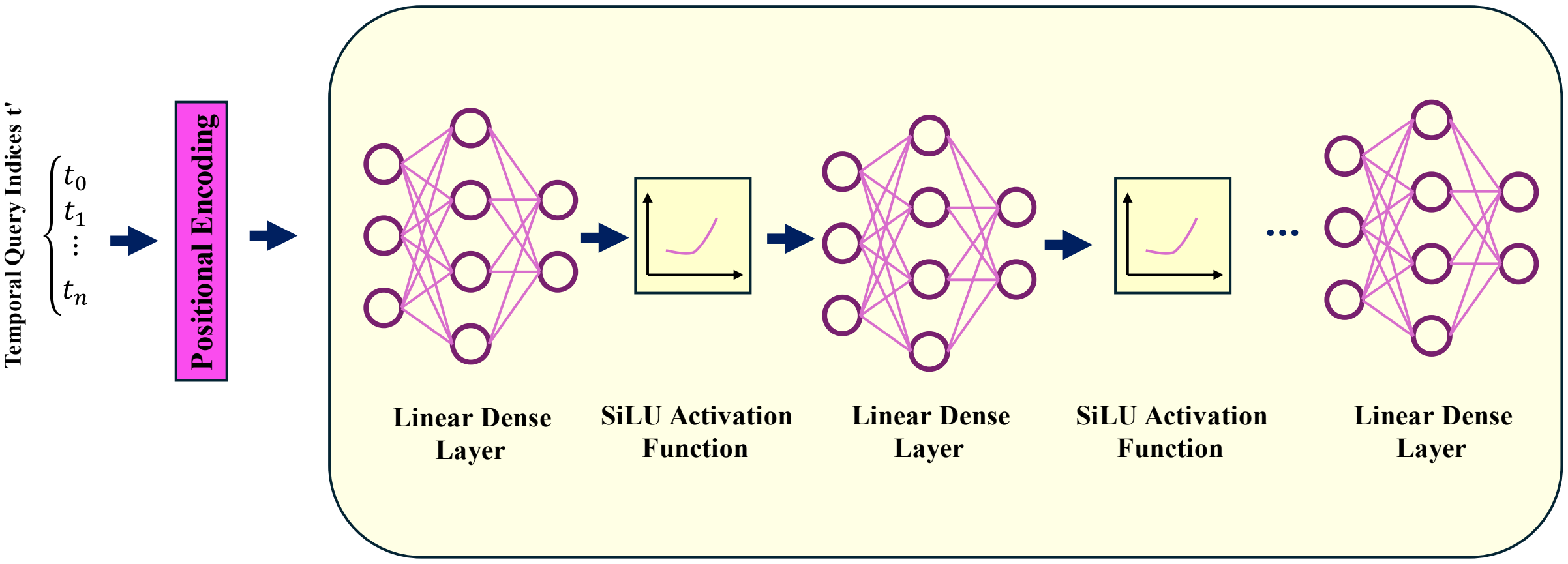

Fig. 3. Architecture of the Trunk Network. The trunk network is a Multi-Layer Perceptron (MLP) responsible for generating time-dependent basis functions. It takes temporal query indices as input, which are first converted into continuous vector representations via Positional Encoding. These encoded vectors are then processed through a series of linear layers with SiLU activation functions to produce a unique feature vector that captures the temporal signature for each corresponding forecast step.

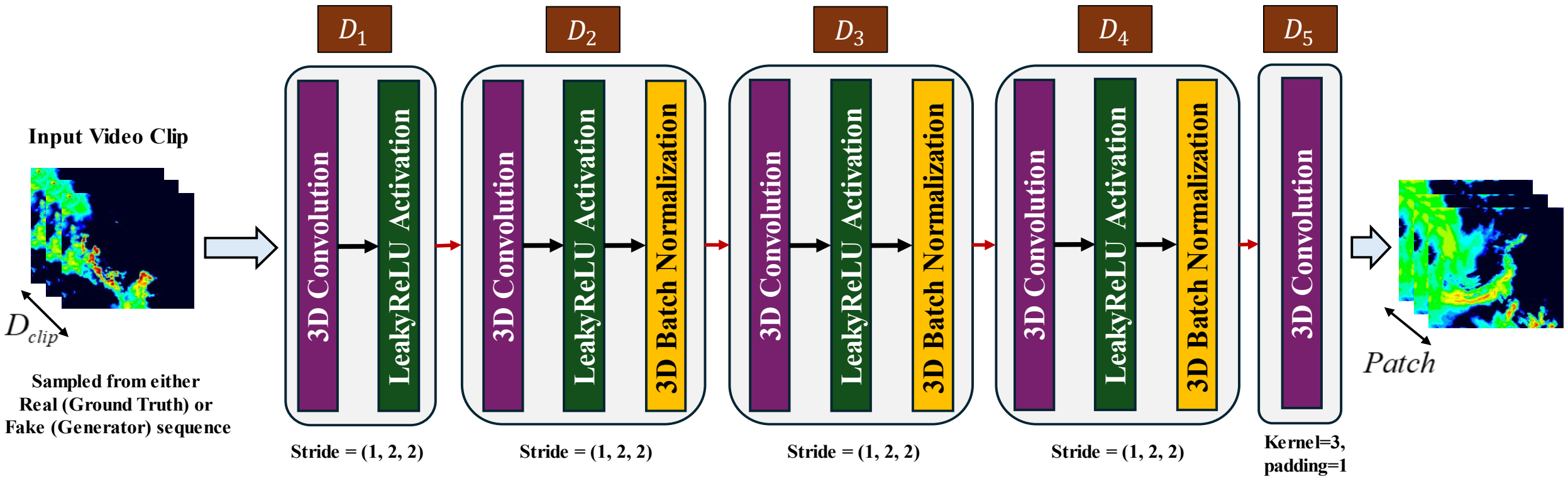

Fig. 4. Architecture of the Spatio-Temporal Discriminator. The discriminator is a deep 3D CNN designed to distinguish between real and generated precipitation sequences. It processes short video clips (sampled from the ground truth or generator's output) through a series of 3D convolutional blocks that downsample the spatial dimensions while preserving temporal information. Operating as a Patch GAN, it outputs a 3D grid of scores, evaluating the authenticity of local spatio-temporal patches rather than a single score for the entire clip. This encourages the generator to produce forecasts that are both spatially realistic within individual frames and temporally coherent across the sequence.

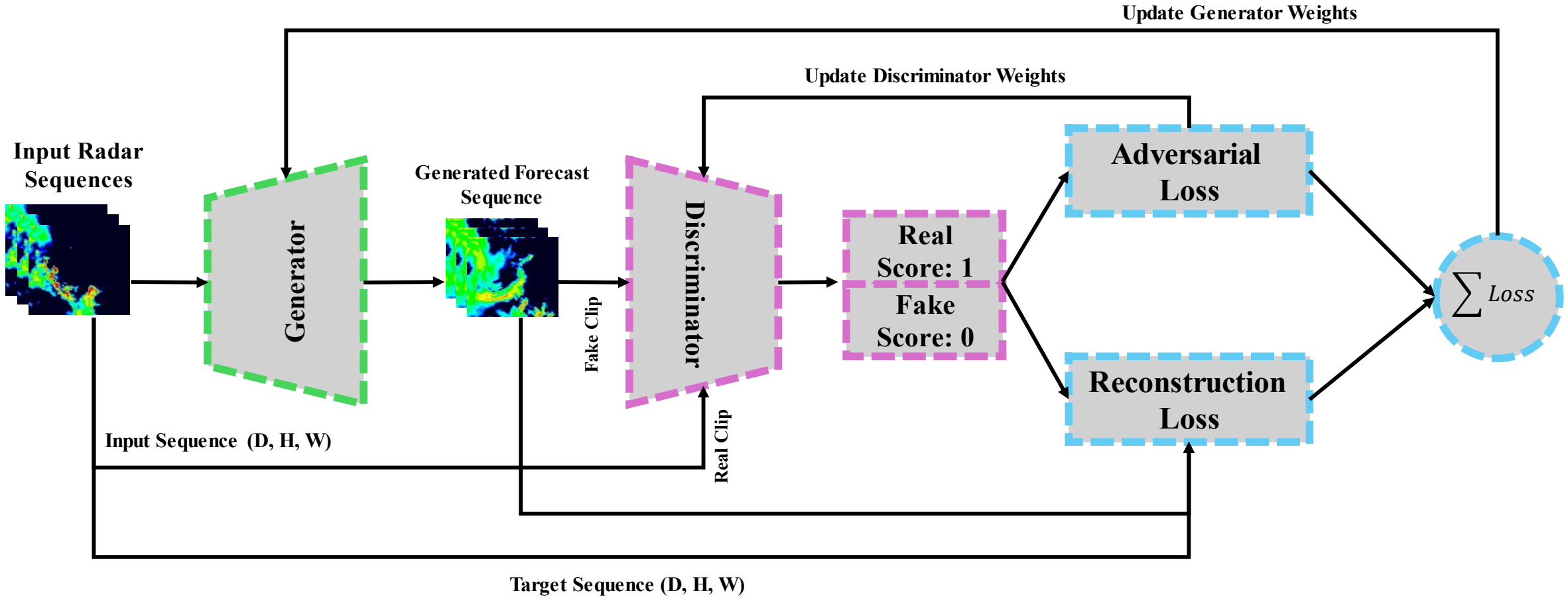

Fig. 5. Overall framework of the GenONet training process. The Generator produces a forecast sequence based on past radar observations. The Discriminator is trained to distinguish these generated sequences from real ground truth sequences. The Generator is then updated based on a composite loss: (1) an adversarial loss that encourages it to 'fool' the discriminator, and (2) a reconstruction loss that ensures pixel-wise and structural similarity to the ground truth. This adversarial loop guides the generator to produce sharper and more realistic forecasts.

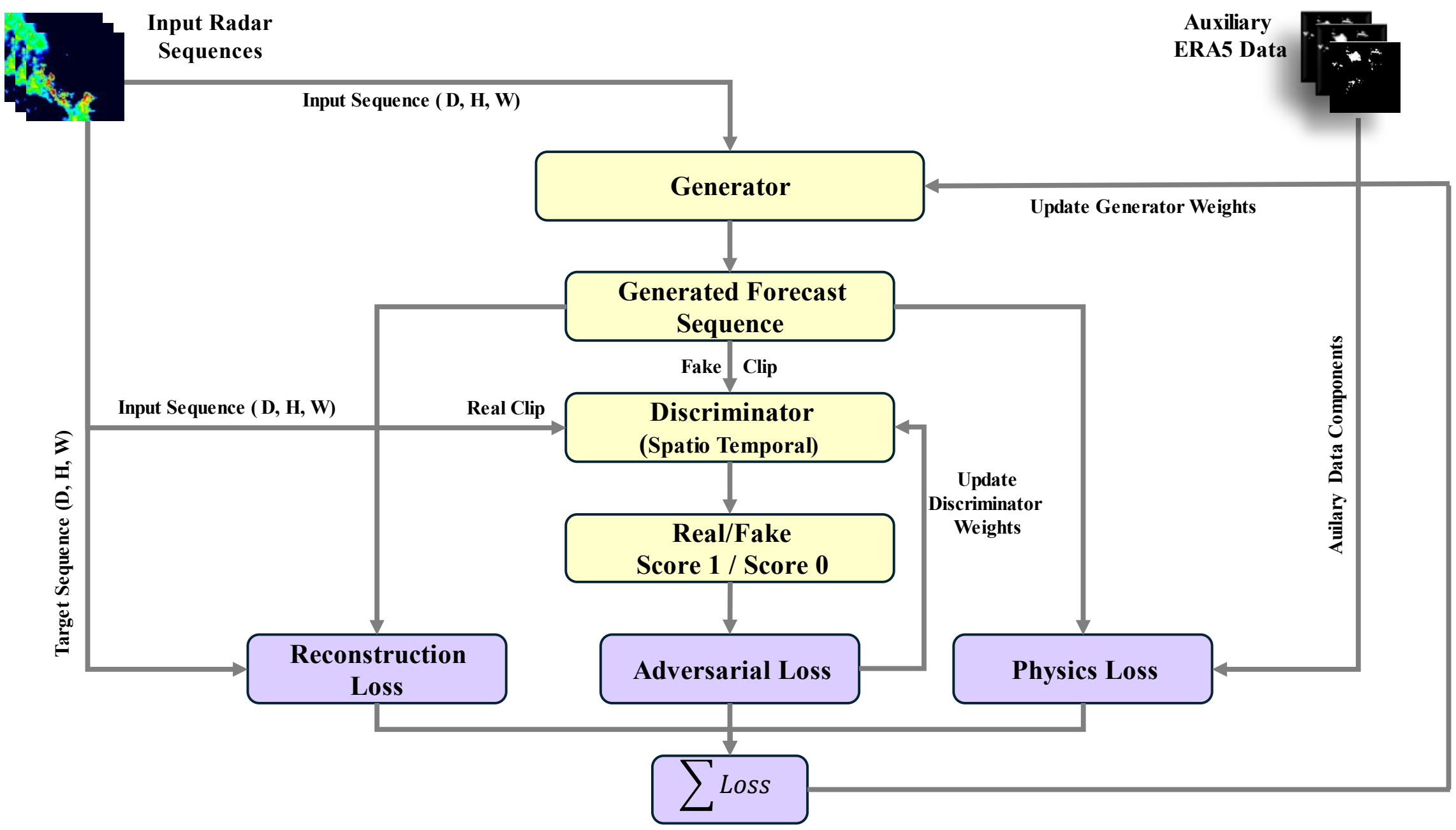

Fig. 6. Framework for the physics-informed ablation study. This diagram illustrates the integration of the physics-informed loss into the GenONet framework. In addition to the adversarial and reconstruction losses, a third loss term is calculated by evaluating the generator's forecast against a simplified Moisture Conservation Equation (MCE), which is informed by auxiliary meteorological data from ERA5. The purpose of this Physics Loss is to act as a regularizer, guiding the generator towards solutions that are not only data-consistent but also more physically plausible.

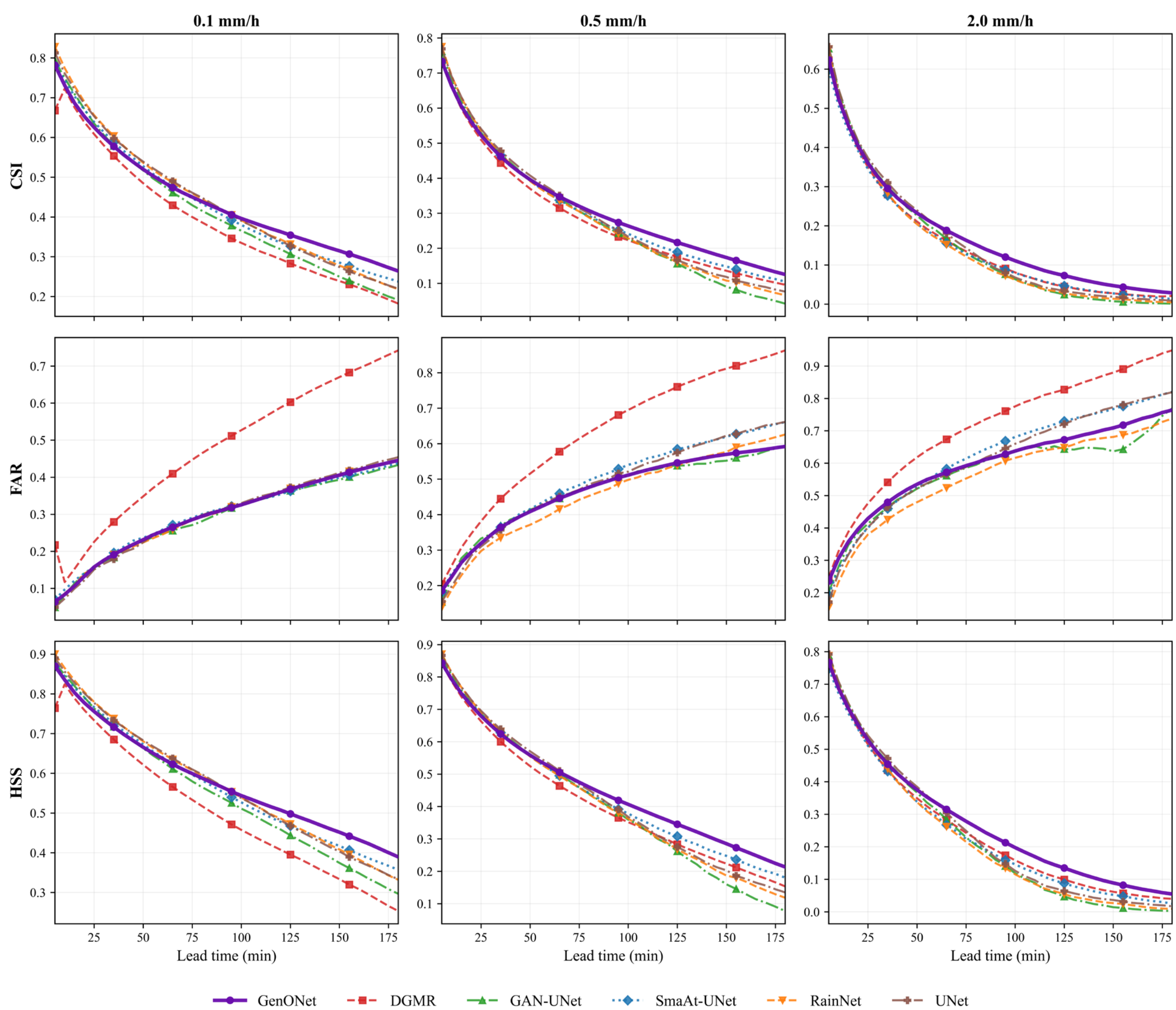

Fig. 7. Comparison of categorical skill scores for GenONet and baseline models as a function of forecast lead time. The metrics shown are CSI, FAR, and HSS evaluated at three rainfall intensity thresholds (0.1, 0.5, and 2 mm/h). GenONet is competitive at short lead times and increasingly outperforms the baselines at longer lead times, where its CSI and HSS advantage widens, particularly for higher intensities. These results demonstrate GenONet's robustness for long-range forecasting of significant weather events.

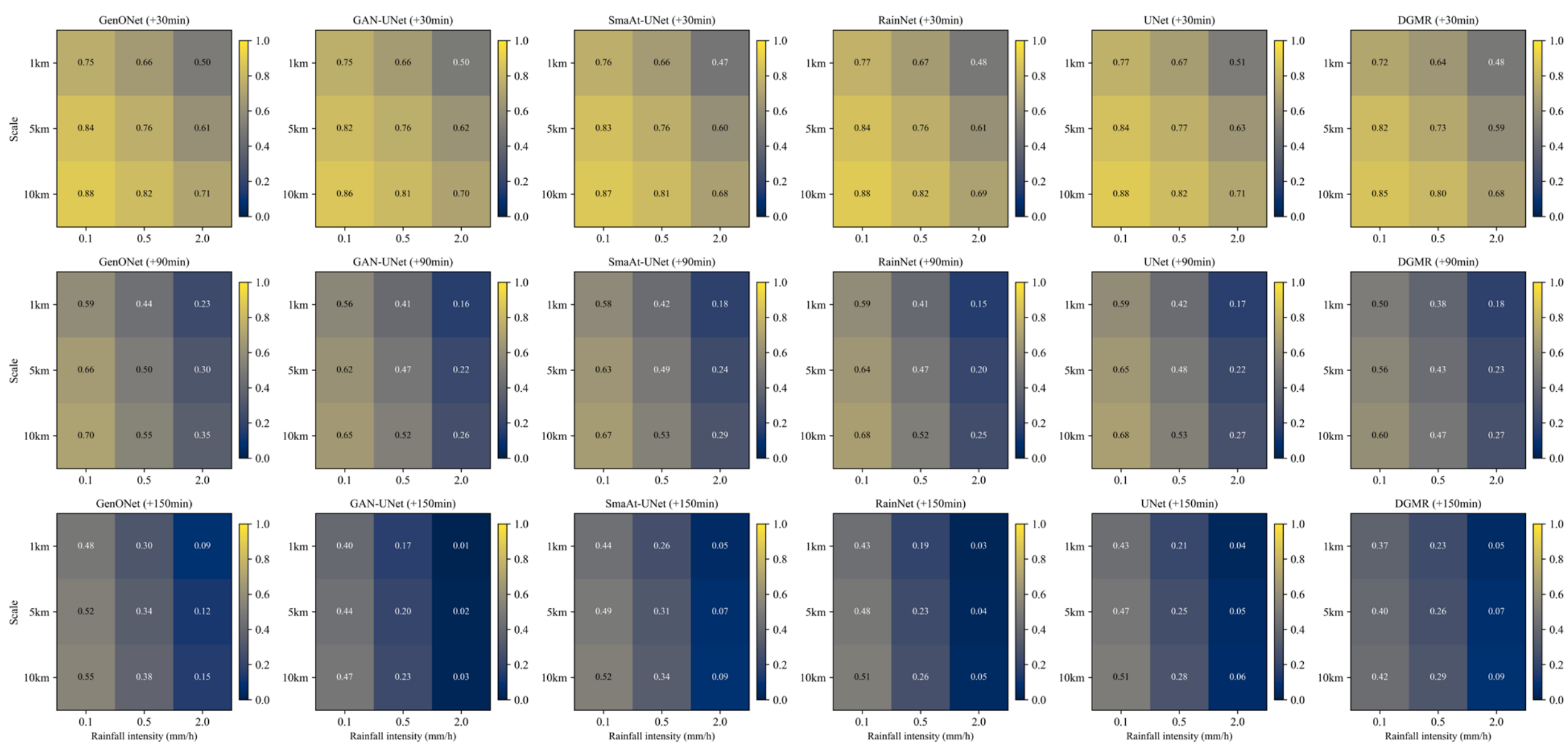

Fig. 8. Heatmap comparison of the FSS at three lead times (+30, +90, and +150 minutes). Each heatmap shows the FSS for a given model across different rainfall intensities (x-axis) and spatial neighborhood sizes (y-axis). Brighter yellow cells indicate higher skill. GenONet is competitive with the baselines at short lead times and shows its clearest advantage at longer lead times and higher intensities, where it retains more spatial skill than the other models as their FSS declines toward zero.

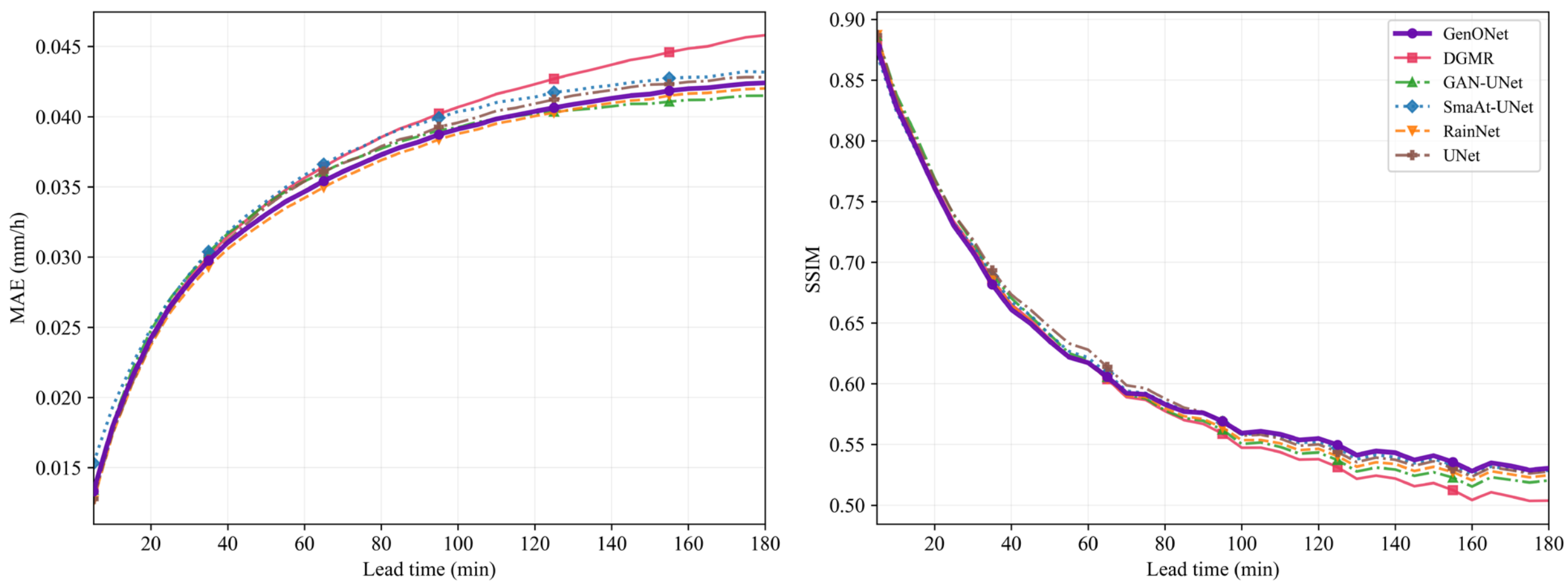

Fig. 9. Comparison of MAE (left) and SSIM (right) over the 180-minute forecast horizon for all models, including DGMR. Lower MAE and higher SSIM indicate better performance. The models perform comparably on MAE, which stays tightly grouped across the horizon. On SSIM, GenONet retains the highest structural similarity at longer lead times, indicating that it preserves the spatial structure and texture of precipitation fields better than the baselines as the forecast horizon increases. Both metrics are computed on physical rainfall fields in mm/h for the primary task ($\Delta T$ = 5 min), averaged over all grid points including dry pixels. Absolute MAE values are therefore small and are not directly comparable with the secondary-task values reported in Table IV.

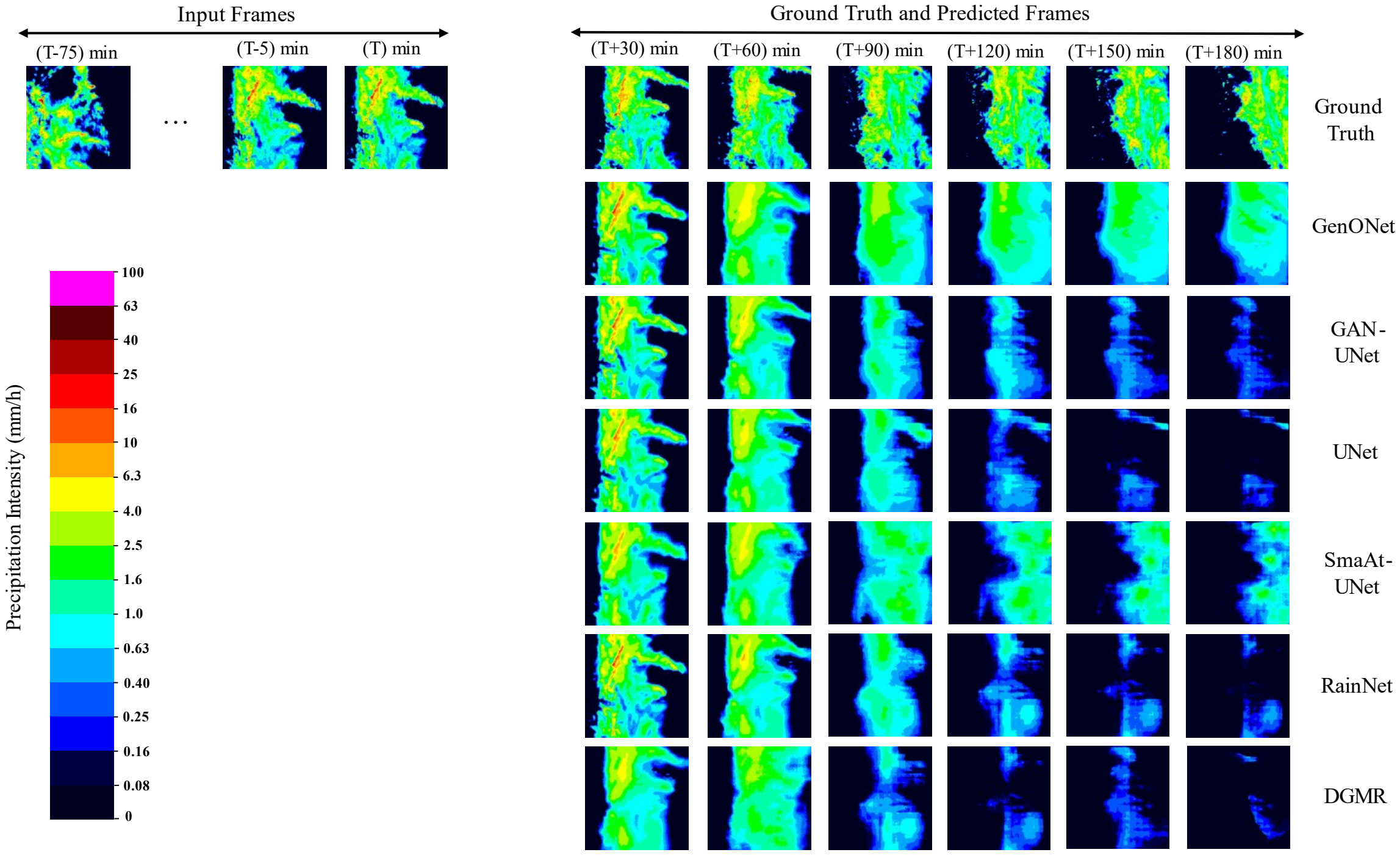

Fig. 10. Qualitative comparison of forecasts from all models for a representative case study. The first three columns show input frames (T-75, T-5, T); the remaining columns show ground truth and predicted precipitation at T+30 to T+180 min. Baseline models progressively lose structural detail and blur at longer lead times, while GenONet better preserves the shape, location, and intensity of the precipitation field across the full horizon.

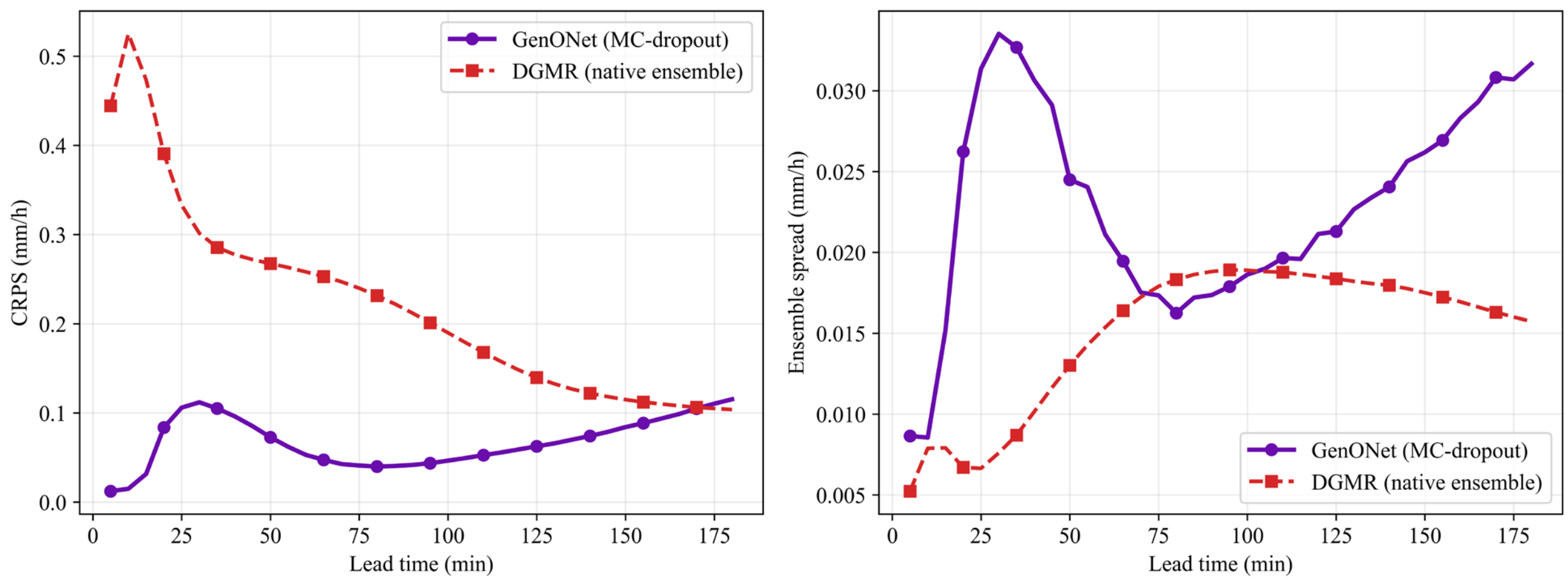

Fig. 11. Probabilistic verification over the 3-hour horizon. CRPS (left) and ensemble spread (right) for GenONet with Monte Carlo dropout and DGMR with its native ensemble, both using 10 members over the same test sequences and reported in mm/h. GenONet attains a mean CRPS of 0.0677 versus 0.221 for DGMR. The convergence of the two CRPS curves near the end of the horizon reflects the weakening of DGMR's forecasts toward near-zero fields rather than improving skill, as indicated by its small ensemble spread.